\documentclass[10pt,twocolumn,letterpaper]{article}

\usepackage{cvpr}        

\usepackage{graphicx}
\usepackage{amsmath}
\usepackage{amssymb}
\usepackage{booktabs}
\usepackage{subcaption}
\usepackage[accsupp]{axessibility}

\usepackage[pagebackref,breaklinks,colorlinks]{hyperref}

\usepackage[capitalize]{cleveref}
\crefname{section}{Sec.}{Secs.}
\Crefname{section}{Section}{Sections}
\Crefname{table}{Table}{Tables}
\crefname{table}{Tab.}{Tabs.}

\usepackage{xcolor}         
\usepackage{multirow}

\def\cvprPaperID{*****} 
\def\confName{CVPR}
\def\confYear{2023}

\begin{document}

\title{WEDGE: A multi-weather autonomous driving dataset built from generative vision-language models}

\author{Aboli Marathe$^1$,  Deva Ramanan$^2$,  Rahee Walambe$^{3,4}$,  Ketan Kotecha$^{3,4}$ \\ \\ 
$^1$Machine Learning Department, Carnegie Mellon University, PA\\
$^2$ Robotics Institute,  Carnegie Mellon University, PA\\
$^3$ Symbiosis Centre for Applied AI (SCAAI), Symbiosis International University (SIU), India\\ 
$^4$Symbiosis Institute of Technology (SIT), Symbiosis International University (SIU), India\\
{\tt\small abolim@cs.cmu.edu, deva@cs.cmu.edu, rahee.walambe@sitpune.edu.in, director@sitpune.edu.in}
}
\maketitle

\begin{abstract}

   The open road poses many challenges to autonomous perception, including poor visibility from extreme weather conditions. Models trained on good-weather datasets frequently fail at detection in these out-of-distribution settings. To aid adversarial robustness in perception, we introduce WEDGE (WEather images by DALL-E GEneration): a synthetic dataset generated with a vision-language generative model via prompting. WEDGE consists of 3360 images in 16 extreme weather conditions manually annotated with 16513 bounding boxes, supporting research in the tasks of weather classification and 2D object detection. We have analyzed WEDGE from research standpoints, verifying its effectiveness for extreme-weather autonomous perception. We establish baseline performance for classification and detection with 53.87\% test accuracy and 45.41 mAP. Most importantly, WEDGE can be used to fine-tune state-of-the-art detectors, improving SOTA performance on {\bf real-world} weather benchmarks (such as DAWN) by {\bf 4.48 AP for well-generated classes like trucks}. WEDGE has been collected under OpenAI's terms\footnote{https://openai.com/policies/terms-of-use. Accessed: March 21, 2023.} of use and is released for public use under the CC BY-NC-SA 4.0 license. The repository for this work and dataset is available at \href{https://infernolia.github.io/WEDGE}{https://infernolia.github.io/WEDGE}.
\end{abstract}

\section{Introduction}

\begin{figure}[h!]
    \centering
    \includegraphics[width=\linewidth,height=8cm]{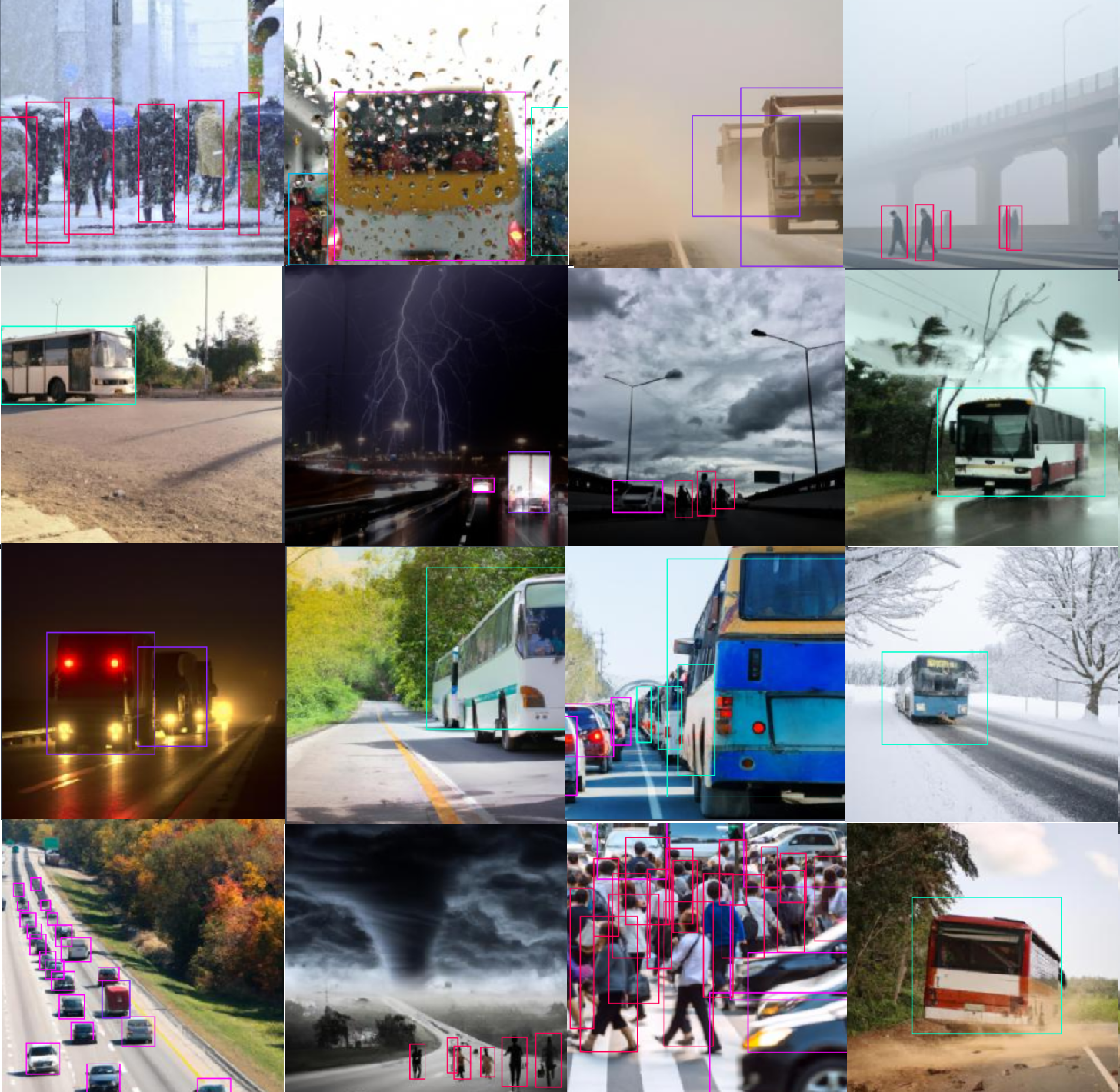}
    \caption{WEDGE synthetic images are generated from vision-language models using prompts of the form ``\{Objects\} on  \{scenes\} when \{weather condition\}". Crucially, weather conditions vary across \{{\tt snowing, raining, dusty, foggy, sunny, lightning, cloudy, hurricane, night, summer, spring, winter, fall, tornado, day, windy}\}, as shown from the top-left to the bottom-right. By fine-tuning detectors on such images manually annotated with bounding boxes, we improve SOTA performance on real-world weather datasets\protect\cite{kenk2020dawn} by \bf 4.48 AP for well-generated classes like trucks.}
    \label{fig10}
\end{figure}

Self-driving cars need to safely operate across diverse weather conditions, generating a demand for extreme-weather perception data.   
This data is mostly captured through fleet operations which are dependent on several factors like sensor calibration, vehicle availability, road condition and equipment costs.  Because of the low-frequency of naturally-encountered adverse weather, manual data collection can be expensive. Moreover, such collection can also be unsafe for extreme weather conditions that reduce visibility or impair vehicle control, such as dust, snow, and fog. Because of such difficulty in data collection, many approaches treat weather conditions (such as rain droplets) as artifacts that can be removed through denoising  \protect\cite{eigen2013restoring,zamir2022restormer,marathe2022restorex}.

One attractive alternative is the use of synthetic data built from rendering engines  \protect\cite{reeves1983particle,sims1990particle}, but such approaches may still not transfer to changing weather conditions or the realism of real-world, due to the so-called {\tt Sim2Real} domain gap and underlying rendering assumptions.
The recent development of realistic synthetic images with generative vision-language models (VLMs) suggests another approach: VLM prompting. We demonstrate that one can use VLMs to build adverse weather datasets for autonomous perception, improving performance on real-world datasets (such as DAWN\protect\cite{kenk2020dawn}) for well-generated classes. Our main contributions include:
\begin{enumerate}
    \item {\bf Data.} First and foremost, we create WEDGE, a 3360 image synthetic dataset of autonomous driving scenes spanning 16 adverse weather conditions. We compare WEDGE to existing datasets, demonstrating that it includes more varied imagery.
    \item {\bf Release.} To allow for public release under the CC BY-NC-SA 4.0 license, we follow guidelines outlined by the VLM's terms of use, manual verifying the quality and appropriateness of the generated images.
    \item {\bf Annotation.} We provide ground truth annotations for all images for two tasks: weather classification and (2D) object detection, with, 16513 bounding box annotations.
    \item {\bf Benchmark.} We establish object detection and classification benchmarks, facilitating future work.
    \item {\bf Sim2Real.} We provide initial evidence that suggests WEDGE can be used for Sim2Real learning; fine-tuning SOTA object detectors on WEDGE improves performance on real-world truck detection by \textbf{4.48} AP. We also examine object classes for which fine-tuning on WEDGE hurts performance. 
\end{enumerate}

The paper is organized as follows. Section \ref{sec:bg} reviews prior datasets. Section \ref{sec:WEDGE} outlines the methodology used to construct and validate WEDGE. Section \ref{sec:exp} presents experimental results for weather classification and Sim2Real object detection. 

\section{Background}
\label{sec:bg}

The relationship between training data and test performance implies better generalization capabilities with better datasets. However, this assessment of “better” datasets can vary based on the respective task, expected performance, distribution requirements and other factors. In the context of autonomous driving tasks, we describe some recent datasets and the general requirements for robust models. We can see that as time progresses, larger datasets also expanded to include more weather conditions for robustness. However, even the best datasets till date do not venture beyond 4 weather types popularly.

Although a number of adverse weather datasets are reported in literature (Refer Table \ref{tab1}), they all pose limitations in two aspects : 1. The data contains the images corresponding to a very few (maximum four) adverse weather scenarios. 2. The data size is small, and it is biased towards a certain city or region and has an inter-class imbalance. When the models trained on these datasets are deployed for real-world weather computer vision tasks, their performance drops significantly in novel weather settings due to lack of heterogeneity and variability. 

Hence, in this work, we report a new dataset which is developed using the DALL-E framework and offers balanced data generated for 16 weather scenarios and multiple object classes. The data is balanced for the all the weather events (210 images per weather class). The object class balance can also be achieved by weighting and re-sampling. Additionally, as the data is developed using generative AI, it is ideally more robust in nature. Some recent works have showcased favorable results using DALL-E and diffusion models for applications including zero-shot classification \protect\cite{li2023your}, detection \protect\cite{ge2022dall} and face generation \protect\cite{borji2022generated}. We provide a number of experimental results in support of robustness and evaluate the usability of this dataset as a benchmarking tool in autonomous perception.

\begin{table*}[h]
\begin{center}
\resizebox{1.95\columnwidth}{!}{\begin{tabular}{llllll} \hline
\textbf{Work} & \textbf{Contribution} & \textbf{Features} & \textbf{\begin{tabular}[c]{@{}l@{}}Class \\ Evaluated\\ /Proposed\end{tabular}} & \textbf{Cities} & \textbf{Weather Condition (S)} \\ \hline
\begin{tabular}[c]{@{}l@{}}KITTI\\ 2012 \protect\cite{geiger2012we}\end{tabular} & \begin{tabular}[c]{@{}l@{}}3D detection, stereo, \\ optical flow, visual \\ odometry/SLAM\end{tabular} & \begin{tabular}[c]{@{}l@{}}22 scenes, stereo data,\\  dense point clouds\end{tabular} & 3/3 & 1 & Good weather only \\
\begin{tabular}[c]{@{}l@{}}CityScapes 2016\protect\cite{cordts2016cityscapes}\end{tabular} & \begin{tabular}[c]{@{}l@{}}2D detection,\\ semantic labeling\end{tabular} & 25000 images & 19/30 & 50 & Good weather only \\
\begin{tabular}[c]{@{}l@{}}Foggy Cityscapes \\Driving 2018 \protect\cite{sakaridis2018semantic}\end{tabular} & \begin{tabular}[c]{@{}l@{}}2D detection,\\  semantic labeling\end{tabular} & 20,550 images & 19/30 & 50 & Fog \\
\begin{tabular}[c]{@{}l@{}}Waymo Open\\ 2020\protect\cite{sun2020scalability}\end{tabular} & \begin{tabular}[c]{@{}l@{}}2D, 3D detection \\ and tracking tasks\end{tabular} & \begin{tabular}[c]{@{}l@{}}1150 scenes, LiDAR \end{tabular} & 4/4 & 3 & \begin{tabular}[c]{@{}l@{}}Good weather \\ with night, rain \end{tabular} \\
\begin{tabular}[c]{@{}l@{}}nuScenes 2020\protect\cite{caesar2020nuscenes}\end{tabular} & 3D detection, tracking & 1000 scenes, Radar data & 10/23 & 2 & \begin{tabular}[c]{@{}l@{}}Weather conditions\\  (sun, rain and clouds)\end{tabular} \\
\begin{tabular}[c]{@{}l@{}}DAWN \\ 2020 \protect\cite{kenk2020dawn}\end{tabular} & 2D detection & 1000 scenes & 6/6 & - & \begin{tabular}[c]{@{}l@{}}Adverse weather:\\ fog, snow, rain and sand\end{tabular} \\

\begin{tabular}[c]{@{}l@{}}Argoverse 2 2023 \protect\cite{wilson2023argoverse}\end{tabular} & \begin{tabular}[c]{@{}l@{}}3d tracking, \\ motion forecasting\end{tabular} & \begin{tabular}[c]{@{}l@{}}1000 scenes,\\ HD maps\end{tabular} & 26/30 & 6 & \begin{tabular}[c]{@{}l@{}}Weather include\\  (sun, rain, snow)\end{tabular} \\ 
\begin{tabular}[c]{@{}l@{}}WEDGE \end{tabular} & \begin{tabular}[c]{@{}l@{}}2D Detection\end{tabular} & \begin{tabular}[c]{@{}l@{}}3360 scenes\end{tabular} & 5/6 & \begin{tabular}[c]{@{}l@{}}Unknown\\ (variable)\end{tabular} & \begin{tabular}[c]{@{}l@{}}Adverse weather in\\  snowing, raining, dusty,\\ foggy, sunny, lightning, \\cloudy, hurricane, night, \\summer, spring, winter,\\tornado, day, wind,fall\end{tabular} \\ 
\hline
\end{tabular}}
 \caption{Recent datasets in autonomous driving.}
 \label{tab1}
 \end{center}
\end{table*}

\section{Methodology}
\label{sec:WEDGE}

The dataset generation process, prompt formulation and image evaluation techniques are discussed here. The paper employs multiple analysis tools, frameworks, and models \protect\cite{cartucho2018robust,bi,NEURIPS2019_9015,muller2020super,dwyer2022roboflow} to deliver the performance evaluation. 
\begin{figure}[t!]
    \centering
    \includegraphics[width=0.48\textwidth]{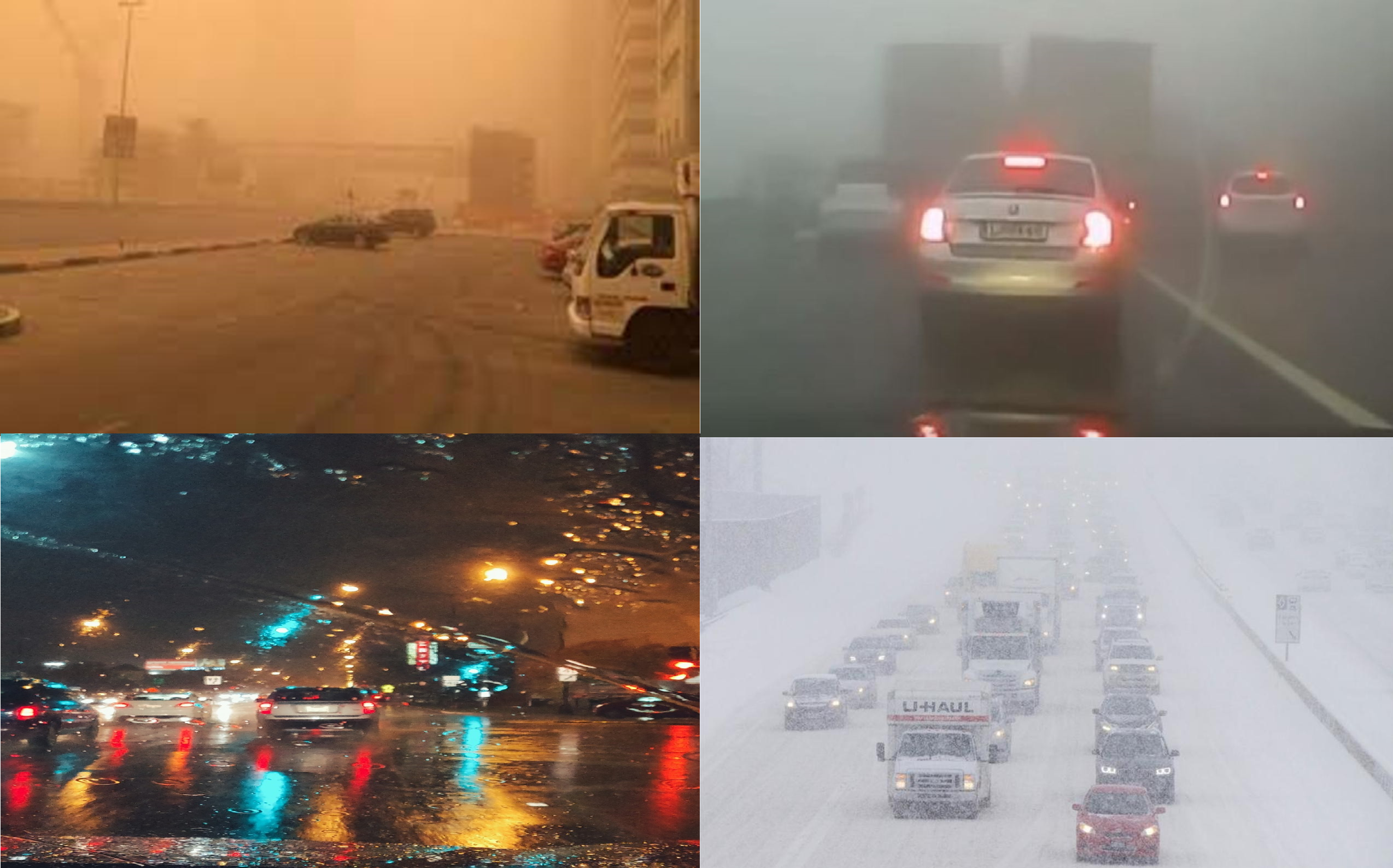}
    \caption{ \textbf{Real dataset samples from DAWN: }Weather conditions in the DAWN dataset \protect\cite{kenk2020dawn} from top left to right: dust, fog,  rainstorm, snowstorm.}
    \label{fig1}
\end{figure}

\subsection{Ground-Truth Dataset}

To test the weather durability of the zero-shot system, we set out to target a range of unfavorable weather situations that can degrade vision in any season. We need a benchmark poor-weather dataset from the actual world for a fair comparison in order to confirm the reliability of this dataset. 

The autonomous vehicle vision dataset: DAWN \protect\cite{kenk2020dawn} with its 1000 driving scenarios recorded in adverse weather conditions was used for this test.
Unfavorable weather conditions that are known to significantly limit road visibility include fog, snow, rain, tornadoes, haze, and sandstorms (Refer Fig \ref{fig1}). Bicycle, person (pedestrian), motorbike, truck, bus, and vehicle (car) form the set of 6 multiscale classes represented in the images.

\subsection{WEDGE Dataset Generation}

The DALL-E \protect\cite{ramesh2021zero} is a large-scale text-to-image generation model that is based on an autoregressive transformer and has shown remarkable generalization capabilities in tasks like zero-shot learning. DALL-E 2  \protect\cite{ramesh2022hierarchical} is a dual-stage model that combines CLIP embeddings with probabilistic diffusion-model based decoder for conditional generation to generate the final realistic images. Diffusion models generate the images based on description (prompt) and sample using this condition. Due to the conditional generation, it presents the opportunity to generate variations in the generated images based on the embeddings. 

\textbf{Data Collection.} OpenAI has provided access to the latest version of DALL-E 2 model through OpenAI API which was used for dataset generation in the following steps:
\begin{enumerate}
    \item 
Collected data using API calls to OpenAI API using prompts which were randomly sampled from the following sets of keywords:

Scenes: highway, road, traffic jam, expressway

Classes: cars, trucks, bus, people crossing

Weather: snowing, raining, dusty, foggy, sunny, lightning, cloudy, hurricane, night, summer, spring, winter, fall, tornado, day, windy
\item Manually verified and cross-examined the images for errors, mismatch and inconsistencies.
\item Grouped images into categories based on weather keywords and thus generated 16 classes with 210 images of each class.
\item Generated 2D bounding box annotations for all images manually using RoboFlow annotation tool \protect\cite{dwyer2022roboflow} and verified with human-in-the-loop evaluation.
\item Explored data using statistical and image analysis techniques, consisting of comparison using image similarity metrics and object-class distribution assessment.
\end{enumerate}
\textbf{Prompt Engineering.} Specifically, we use prompts of the form ``\{Objects\} on  \{scenes\} when \{weather\}", where objects $\in$ \{{\tt cars, trucks, bus, people crossing}\}, scenes $\in$ \{{\tt highway, road, traffic jam, expressway}\}, and weather $\in$ \{{\tt snowing, raining, dusty, foggy, sunny, lightning, cloudy, hurricane, night, summer, spring, winter, fall, tornado, day, windy}\}. This is 4*4= 16 unique prompts for each weather condition, which we randomly queried 210 times to generate a final dataset of 16*210 = 3360 images. For the internal diagnostic analysis presented in Sec.\ref{sec:exp}, we randomly split WEDGE into a 80/20 train/test split for classification.

\subsection{Image Similarity}

We evaluate the threshold differences in image similarity between sampled real and generated images in their respective class clusters, and bin them as shown in Figure \ref{fig6}.
The Information theoretic-based Statistic Similarity Measure combines the statistical method and information theory, and it has a strong ability to forecast the relationship between the image intensity values\protect\cite{aljanabi2019design}.
Peak Signal-to-Noise Ratio (PSNR), which directly operates with image intensity, evaluates the ratio between the maximum possible power of a signal and the power of corrupting noise \protect\cite{hore2010image}. The Root Mean Squared Error (RMSE) calculates the percentage change in each pixel between the operation and the baseline \protect\cite{sara2019image}. Spectral Angle Mapper (SAM) calculates the angle between two spectra and treats them as vectors in a space with a dimensionality equal to the number of bands in order to estimate the spectral similarity between them\protect\cite{yuhas1992discrimination}.Signal to Reconstruction Error Ratio (SRE) is a metric that compares the error to the signal's power\protect\cite{lanaras2018super}. The Structural Similar Index Measure (SSIM) is a tool that aims to capture an image's loss of structure  \protect\cite{hore2010image}.

\section{Experiments}
\label{sec:exp}

\subsection{Image Analysis}

\begin{figure}[h]
    \centering
    \includegraphics[width=0.49\textwidth]{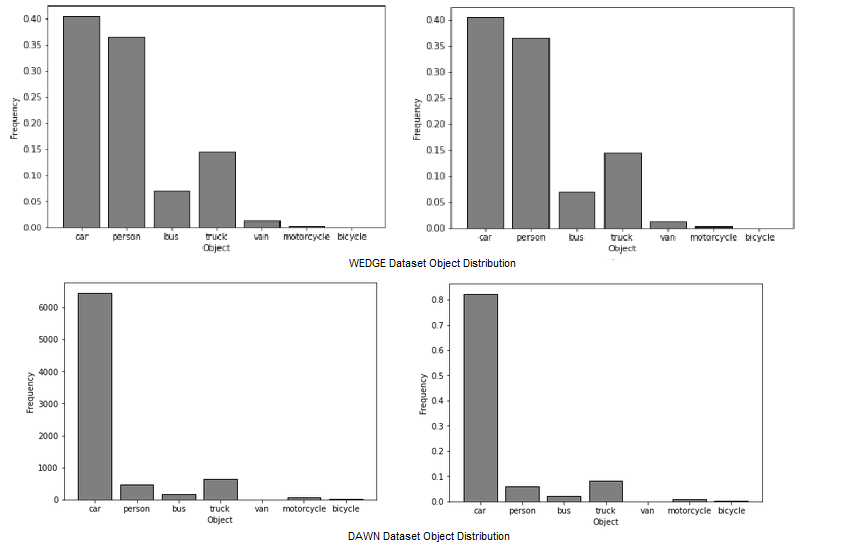}
    \caption{\textbf{Sim2Real Distribution Gap: }Object frequency distribution in WEDGE and DAWN datasets. }
    
    \label{fig2}
\end{figure}

The classic autonomous vehicle settings contain skewed object distributions which we attempt to model with this generated dataset as visible in Figure \ref{fig2}. In practice, this balance can be restored by weighted prompting techniques and resampling if required, but should be maintained to deliver valid results benchmarking generalization capabilities.

\begin{figure}[h]
    \centering
    \includegraphics[width=0.49\textwidth]{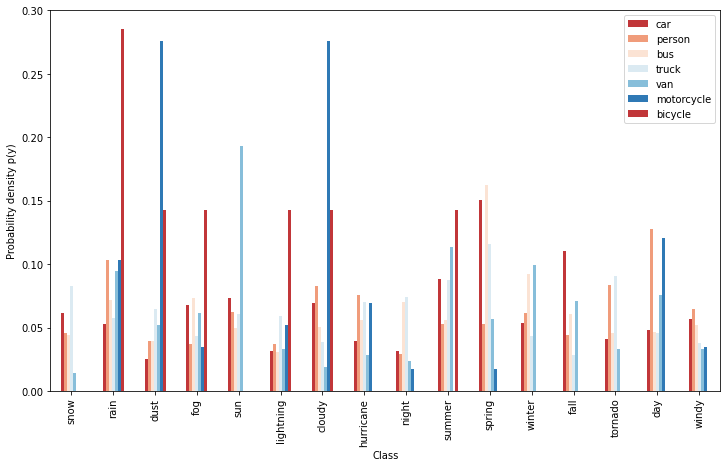}
    \caption{\textbf{Class Imbalance: }Inter-class object distribution in WEDGE dataset .}
    \label{fig3}
\end{figure}

We observe that the inter-class object distribution is also unbalanced (Figure \ref{fig3}), which is a desirable quantity while training for robustness. In the wild, autonomous driving scenes will present unbalanced object distributions, which are difficult to perceive with detectors trained on fairly balanced data \protect\cite{oksuz2020imbalance}.

\subsection{Image Similarity Analysis}

\begin{figure}[h]
    \centering
  
        \includegraphics[width=0.5\textwidth,height=15.85cm]{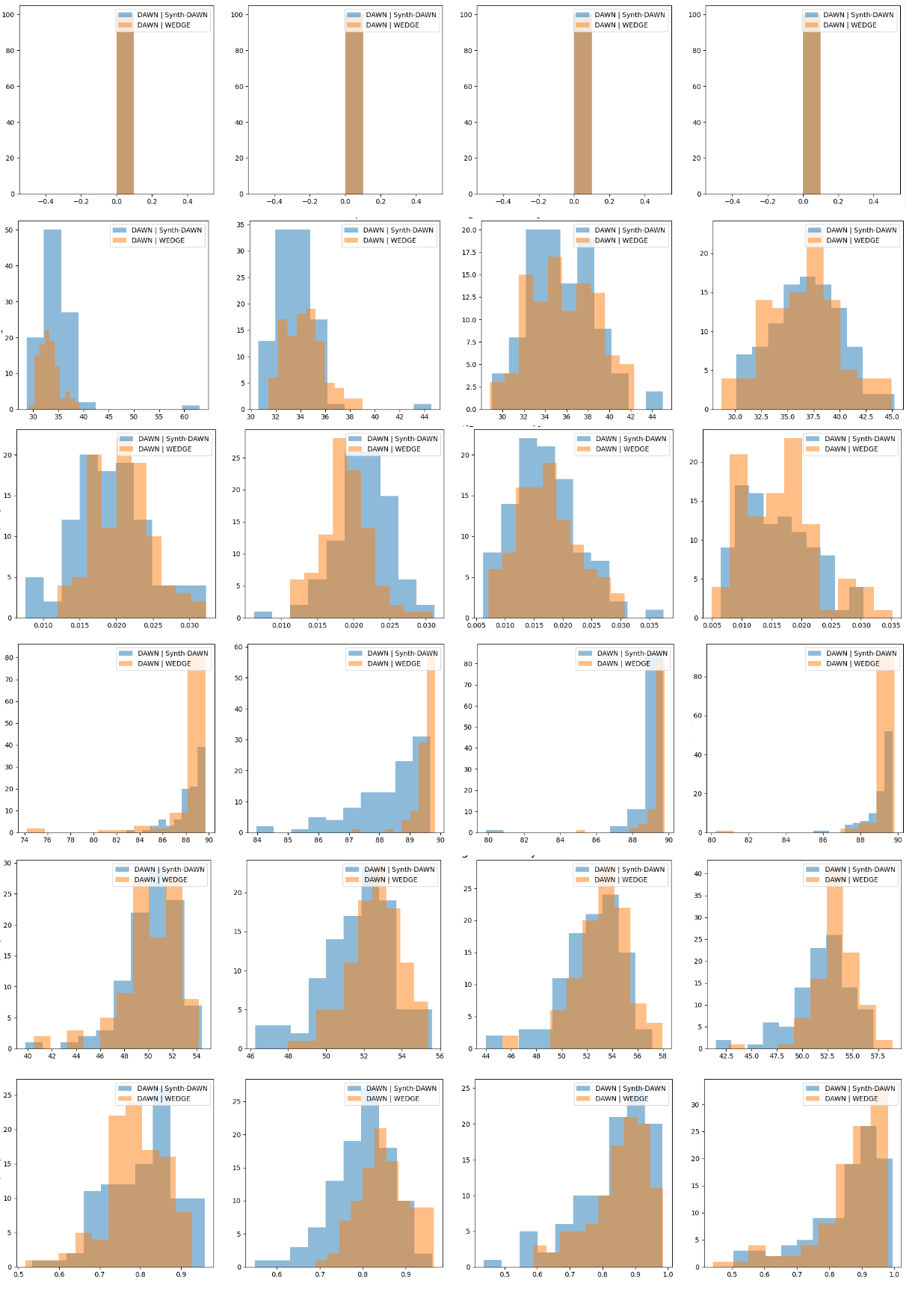}
        \caption{\textbf{Image similarity thresholds of modified real (Synth-DAWN) and synthetic autonomous driving datasets with real images as evaluated using 6 metrics (ISSM, PSNR, RMSE, SAM, SRE, SSIM from top to bottom) indicate overlapping similarity distributions}. The Sim2Real gap as evaluated by these metrics is comparable to Filter2Real gap of applying  simple filters (like blurring, sharp edges without distorting the structural similarities). On the y-axis is the frequency of binned similarity(error) and x-axis is the bins of similarity. Orange color represents similarity between a sampled image from DAWN and WEDGE (Sim2Real). Blue color is similarity shift Filter2Real from common filtering modifications between a sampled image from DAWN and a modified sampled image from DAWN which is called Synth-DAWN. }
    \label{fig6}

\end{figure}

We evaluate the real and generated datasets side by side using these 6 metrics and as seen in the figure \ref{fig6}, we hypothesize a sensible range of errors in  this relative difference between real and generated datasets. 
The expected inverse similarity should ideally bounded by a small valued real number which varies according to the properties of the similarity metric. 

\section{Results}

\begin{table}[h]
\resizebox{\columnwidth}{!}{
\begin{tabular}{lll} \hline
\textbf{Model} & \textbf{Train Acc.} & \textbf{Test Acc.}  \\ \hline
VGG16 \protect\cite{simonyan2014very}& 95.80 & 44.35\\
VGG19 \protect\cite{simonyan2014very}& 98.59 & 44.64  \\
Xception \protect\cite{chollet2017xception}& 98.33 & 46.73  \\
ResNet50 \protect\cite{he2016deep}& 35.04 & 22.47  \\
ConvNeXtSmall\protect\cite{liu2022convnet}& 58.97 & 18.15 \\
InceptionV3 \protect\cite{szegedy2016rethinking}& 99.85 & 50.30 \\
MobileNet\protect\cite{howard2017mobilenets}& 99.33 & \textbf{53.87}  \\
MobileNetv2\protect\cite{howard2017mobilenets}& 99.67 & 46.43 \\
DenseNet\protect\cite{huang2017densely}& \textbf{99.89} & 49.55  \\
EfficientNetV2S \protect\cite{tan2021efficientnetv2}& 35.01 & 18.75 \\ \hline
\end{tabular}
}
 \caption{\textbf{Weather Classification}: Classifying the weather condition of a WEDGE image, in supervised settings with a 80/20 train/test split across 10 selected models. We see that the models can predict weather conditions with reasonable accuracy. Random classification obtains 6.25\% accuracy, which can be used to compare obtained results. The MobileNet \protect\cite{howard2017mobilenets} Classifier achieves top performance on the WEDGE Dataset with 53.87\% test accuracy which is over 8-fold improvement on random classification.}
 \label{tab2}
\end{table}

\begin{table*}[h]
\begin{center}

\resizebox{1.9\columnwidth}{!}{\begin{tabular}{lllllllllllllll}
\hline
\multicolumn{1}{l|}{} & \multicolumn{8}{l|}{\textbf{Real Data (DAWN Dataset)}} & \multicolumn{6}{l}{\textbf{Synthetic Data (WEDGE Dataset)}} \\ \cline{2-15} 
\multicolumn{1}{l|}{\multirow{-2}{*}{\textbf{Model}}} & \textbf{car} & \textbf{person} & \textbf{bus} & \textbf{truck} & \textbf{T-4 AP} & \textbf{mc} & \textbf{bicycle} & \multicolumn{1}{l|}{\textbf{mAP}} & \textbf{car} & \textbf{person} & \textbf{bus} & \textbf{truck} & \textbf{van} & \textbf{mAP} \\ \hline
\multicolumn{15}{l}{\textbf{Prior Art}} \\ \hline
\multicolumn{1}{l|}{Multi-weather city \protect\cite{ms}} & - & - & - & - & 21.20 (39.19) & - & - & \multicolumn{1}{l|}{(39.19)} & - & - & - & - & - & - \\
\multicolumn{1}{l|}{RoHL \protect\cite{saikia2021improving}} & - & - & - & - & - & - & - & \multicolumn{1}{l|}{28.80} & - & - & - & - & - & - \\
\multicolumn{1}{l|}{Transfer Learning \protect\cite{marathe2022rain}} & 7.00 & 8.00 & 7.00 & - & 5.50 & - & 0.00 &  \multicolumn{1}{l|}{-} & - & - & - & - & - & - \\
\multicolumn{1}{l|}{Data Augmentation \protect\cite{marathe2022rain}} & 6.00 & 4.00 & 3.00 & 0.00 & 26.25 & - & \textbf{92.00} & \multicolumn{1}{l|}{-} & - & - & - & - & - & - \\
\multicolumn{1}{l|}{\begin{tabular}[c]{@{}l@{}}Weather-\\ Night GAN \protect\cite{marathe2022restorex}\end{tabular}} & 48.00 & 0.00 & 0.00 & 0.00 & 12.00 & - & - & \multicolumn{1}{l|}{-} & - & - & - & - & - & - \\
\multicolumn{1}{l|}{Ensemble Detectors \protect\cite{a3}} & 52.56 & 52.34 & 21.73 & 13.71 & 35.08 & 35.51 & 23.29 & \multicolumn{1}{l|}{32.75} & - & - & - & - & - & - \\ \hline \hline
\multicolumn{15}{l}{\textbf{Evaluation on DAWN-All}} \\ \hline
\multicolumn{15}{l}{\textbf{Trained on Good Weather Data (COCO \protect\cite{lin2014microsoft})}} \\ \hline

\multicolumn{1}{l|}{\begin{tabular}[c]{@{}l@{}}FasterRCNN   \\ MobileNet \\ Large 320 \protect\cite{ren2015faster,howard2017mobilenets}\end{tabular}} &37.56	&	34.93	&	20.90	&	12.91	&	26.57	&	23.15	&	18.95	&	\multicolumn{1}{l|}{24.73} & 34.10	&	36.26	&	39.35	&	16.05	&	0.00	&	25.15\\
\multicolumn{1}{l|}{\begin{tabular}[c]{@{}l@{}}FasterRCNN   \\ MobileNet\\ Large \protect\cite{ren2015faster,howard2017mobilenets}\end{tabular}} & 60.64	&	55.96	&	32.78	&	23.66	&	43.26	&	38.55	&	28.75	&	\multicolumn{1}{l|}{40.05} & 35.34	&	39.52	&	35.83	&	25.43	&	0.00	&	27.22\\
\multicolumn{1}{l|}{FasterRCNN ResNet 50\protect\cite{ren2015faster}} &  \textbf{69.13}	&	\textbf{70.31}	&	\textbf{38.64}	&	30.54	&	\textbf{52.15}	&	\textbf{52.17}	&	\textbf{30.56}	&	\multicolumn{1}{l|}{\textbf{48.55}}  & 31.41	&	33.54	&	30.19	&	18.75	&	0.00	&	22.78 \\ \hline

\multicolumn{15}{l}{\textbf{Fine-Tuning on WEDGE}} \\ \hline
\multicolumn{1}{l|}{\begin{tabular}[c]{@{}l@{}}FasterRCNN   \\ MobileNet \\ Large 320 \protect\cite{ren2015faster,howard2017mobilenets}\end{tabular}} & \textbf{39.52}	&	23.97	&	7.81	&	\textbf{22.08}	&	23.34	&	0.00	&	0.00	&	\multicolumn{1}{l|}{15.56} & 40.40	&	43.01	&	49.88	&	31.41	&	10.19	&	34.98 \\
\multicolumn{1}{l|}{\begin{tabular}[c]{@{}l@{}}FasterRCNN   \\ MobileNet \\ Large \protect\cite{ren2015faster,howard2017mobilenets}\end{tabular}} & 59.81	&	34.61	&	14.06	&	\textbf{30.67}	&	34.78	&	0.00	&	0.00	&	\multicolumn{1}{l|}{23.19}  & 52.52	&	\textbf{54.79}	&	\textbf{51.23}	&	50.01	&	7.95	&	43.30\\ 
\multicolumn{1}{l|}{FasterRCNN ResNet 50\protect\cite{ren2015faster}} &  68.09	&	54.29	&	27.48	&	\textbf{35.02}	&	46.22	&	0.00	&	0.00	&	\multicolumn{1}{l|}{30.81} & \textbf{57.48}	&	54.71	&	46.92	&	\textbf{57.43}	&	\textbf{10.49}	&	\textbf{45.41}\\ \hline
\end{tabular}}

\caption{\textbf{Object Detection: } Performance for {\tt Car, Person, Bus, Truck, Van, Motorcycle (mc), Bicycle} using the PASCAL VOC mAP metric on real (DAWN) and our synthetic (WEDGE) data. Previous work uses different protocols for evaluation on DAWN;  \protect\cite{ms} evaluates on DAWN WD set (Fake droplets on fake wet generated conditions) and reports the overall AP averaged over classes (AP@50 is included in brackets: Improvement over this value is 12.96 AP on DAWN-All and 18.86 AP on DAWN-Test), \protect\cite{saikia2021improving} evaluates on corrupted testsets and reports average AP across corruptions, \protect\cite{marathe2022rain,marathe2022restorex} evaluates on 1000 random images, while \protect\cite{a3} evaluates on 500 random images of DAWN and reports AP and mAP. DAWN has a 90-10 train-test split (proposed), but since our models are not trained on DAWN, we present results for DAWN-All (and include results for DAWN-Test in Table \ref{tab6}). 
First, we find that simply evaluating state-of-the-art (SOTA) {\em off-the-shelf} (OTS) object detectors (trained on {\em good} weather data) already outperforms all published results. This establishes our pre-trained detectors as strong baselines for this task. Fine-tuning such models (specifically, ResNet50) on WEDGE further improves truck AP by 4.48 on DAWN-All (4.44 AP on DAWN-Test). The fine-tuned MobileNet-Large is able to detect both cars and trucks better with 1.96 AP and 9.17 AP on DAWN-All and (2.61 AP and 5.17 AP on DAWN-Test) respectively. T-4 AP is the averaged AP over 4 key object classes {\tt Car, Person, Bus, Truck}. In general, fine-tuning other categories tend to hurt performance, which we discuss further in the text.\\
*Additionally, \protect\cite{rs} evaluates on a 3:1 train-test holdout split of DAWN (trained on adverse weather data) and reports overall Vehicle AP. It presents a vehicle detection benchmark on DAWN with 89.48 AP, including comparisons to previous models \cite {cai2018cascade,hu2018sinet,hassaballah2020local,chabot2017deep,xiang2017subcategory,lin2017focal,redmon2018yolov3,li2019scale, he2017mask} in classic supervised learning settings (trained on adverse weather data- DAWN). An analysis of vehicle category results over car, bus and trucks is presented in\cite{hassaballah2020vehicle}. }
 \label{tab3}
     
\end{center}
\end{table*}

\begin{table*}[h]
\begin{center}

\resizebox{1.98\columnwidth}{!}{
\begin{tabular}{llccccccccllllllll}
\hline
\multicolumn{1}{l|}{\textbf{Dataset}} & \multicolumn{1}{l|}{\textbf{}} & \multicolumn{8}{c|}{\textbf{DAWN}} & \multicolumn{8}{c}{\textbf{WEDGE}} \\ \hline
\multicolumn{1}{l|}{\textbf{Model}} & \multicolumn{1}{l|}{\textbf{Weather}} & \textbf{car} & \textbf{person} & \textbf{bus} & \textbf{truck} & \textbf{T-4 AP} & \textbf{mc} & \textbf{bicycle} & \multicolumn{1}{c|}{\textbf{mAP}} & \multicolumn{1}{c}{\textbf{car}} & \multicolumn{1}{c}{\textbf{person}} & \multicolumn{1}{c}{\textbf{bus}} & \multicolumn{1}{c}{\textbf{truck}} & \multicolumn{1}{c}{\textbf{T-4 AP}} & \multicolumn{1}{c}{\textbf{van}} & \multicolumn{1}{c}{\textbf{bicycle}} & \multicolumn{1}{c}{\textbf{mAP}} \\ \hline
\multicolumn{18}{l}{\textbf{Trained on Good   Weather Data (COCO \protect\cite{lin2014microsoft})}} \\ \hline
\multicolumn{1}{l|}{\multirow{4}{*}{\textbf{\begin{tabular}[c]{@{}l@{}}FasterRCNN\\  MobileNet   \\ Large 320\\ \protect\cite{ren2015faster,howard2017mobilenets}\end{tabular}}}} & \multicolumn{1}{l|}{\textbf{Rain}} & 39.61 & 60.64 & 18.66 & 11.5 & 32.60 & 56.67 & - & \multicolumn{1}{c|}{37.41} & 23.31 & 42.56 & 52.42 & \textbf{56.02} & \textbf{43.58} & 0 & \textbf{-} & 34.86 \\
\multicolumn{1}{l|}{} & \multicolumn{1}{l|}{\textbf{Snow}} & 38.48 & 41.25 & 12.47 & 11.43 & 25.90 & \textbf{100} & \textbf{100} & \multicolumn{1}{c|}{43.38} & 41.93 & 26.65 & 33.33 & 42.11 & 36.01 & 0 & - & 28.81 \\
\multicolumn{1}{l|}{} & \multicolumn{1}{l|}{\textbf{Fog}} & 46.43 & 22.12 & 28.12 & 23.6 & 30.06 & 40 & \textbf{100} & \multicolumn{1}{c|}{43.38} & 40.26 & 0 & 67.76 & 40.4 & 37.11 & 0 & 0 & 24.74 \\
\multicolumn{1}{l|}{} & \multicolumn{1}{l|}{\textbf{Dust}} & 34.72 & 30.09 & 24.57 & 14.36 & 25.93 & 20.58 & 14.98 & \multicolumn{1}{c|}{23.21} & \textbf{51.85} & - & 39.77 & 32.73 & 41.45 & \textbf{-} & - & \textbf{41.45} \\
\multicolumn{1}{l|}{\multirow{4}{*}{\textbf{\begin{tabular}[c]{@{}l@{}}FasterRCNN \\ MobileNet   \\ Large\\ \protect\cite{ren2015faster,howard2017mobilenets}\end{tabular}}}} & \multicolumn{1}{l|}{\textbf{Rain}} & 65.35 & 74.95 & 17.85 & 31.59 & 47.43 & 32.14 & - & \multicolumn{1}{c|}{44.38} & 28 & \textbf{43.28} & 41.84 & 55.4 & 42.13 & 0 & - & 33.7 \\
\multicolumn{1}{l|}{} & \multicolumn{1}{l|}{\textbf{Snow}} & 63.25 & 68.32 & 35.66 & 25.41 & 48.16 & 33.33 & \textbf{100} & \multicolumn{1}{c|}{46.57} & 37.67 & 36.63 & 2.94 & 37.64 & 28.72 & 0 & - & 22.98 \\
\multicolumn{1}{l|}{} & \multicolumn{1}{l|}{\textbf{Fog}} & 60.62 & 37.58 & 40.1 & 22.51 & 40.20 & 46.67 & \textbf{100} & \multicolumn{1}{c|}{51.25} & 45.04 & 1.89 & \textbf{71.48} & 48.15 & 41.64 & 0 & 0 & 27.76 \\
\multicolumn{1}{l|}{} & \multicolumn{1}{l|}{\textbf{Dust}} & 56.95 & 49.77 & 34.42 & 21.45 & 40.64 & 38.94 & 26.49 & \multicolumn{1}{c|}{38} & 41.41 & \textbf{-} & 41.01 & 39.38 & 40.6 & - & - & 40.6 \\
\multicolumn{1}{l|}{\multirow{4}{*}{\textbf{\begin{tabular}[c]{@{}l@{}}FasterRCNN \\ ResNet 50\\ \protect\cite{ren2015faster}\end{tabular}}}} & \multicolumn{1}{l|}{\textbf{Rain}} & 71.77 & 70.1 & 21.41 & \textbf{43.8} & 51.77 & 33.33 & - & \multicolumn{1}{c|}{48.08} & 33.31 & 42.24 & 31.3 & 36.19 & 35.76 & 0 & - & 28.61 \\
\multicolumn{1}{l|}{} & \multicolumn{1}{l|}{\textbf{Snow}} & \textbf{72.93} & \textbf{82.69} & \textbf{48.6} & 32.63 & \textbf{59.21} & 25 & \textbf{100} & \multicolumn{1}{c|}{51.69} & 34.71 & 21.7 & 0.31 & 18.59 & 18.83 & 0 & - & 15.06 \\
\multicolumn{1}{l|}{} & \multicolumn{1}{l|}{\textbf{Fog}} & 70.99 & 69.98 & 39.23 & 24.56 & 51.19 & 71.43 & \textbf{100} & \multicolumn{1}{c|}{\textbf{62.7}} & 47.73 & 0 & 55.88 & 40.86 & 36.12 & 0 & 0 & 24.08 \\
\multicolumn{1}{l|}{} & \multicolumn{1}{l|}{\textbf{Dust}} & 65.64 & 64.49 & 43.35 & 25.99 & 49.86 & 54.42 & 26.38 & \multicolumn{1}{c|}{46.71} & 29.71 & - & 3.07 & 26.88 & 19.89 & - & \textbf{-} & 19.89 \\ \hline
\multicolumn{18}{l}{\textbf{Fine-Tuning on   WEDGE}} \\ \hline
\multicolumn{1}{l|}{\multirow{4}{*}{\textbf{\begin{tabular}[c]{@{}l@{}}FasterRCNN \\ MobileNet   \\ Large 320\\ \protect\cite{ren2015faster,howard2017mobilenets}\end{tabular}}}} & \multicolumn{1}{l|}{\textbf{Rain}} & 42.52 & 43.32 & 8.56 & 29.44 & 30.96 & 0 & - & \multicolumn{1}{c|}{24.77} & 42.45 & 42.13 & 68.7 & 81.18 & 58.62 & 0 & - & 46.89 \\
\multicolumn{1}{l|}{} & \multicolumn{1}{l|}{\textbf{Snow}} & 39.44 & 27.06 & 5.81 & 27.63 & 24.98 & 0 & 0 & \multicolumn{1}{c|}{14.28} & 49.82 & 36.78 & 33.33 & 68.62 & 47.14 & 0 & - & 37.71 \\
\multicolumn{1}{l|}{} & \multicolumn{1}{l|}{\textbf{Fog}} & 47.95 & 5.77 & 15.33 & 34.77 & 25.95 & 0 & 0 & \multicolumn{1}{c|}{17.3} & 47.98 & \textbf{100} & 81.87 & 77.1 & 76.74 & \textbf{100} & 0 & 67.82 \\
\multicolumn{1}{l|}{} & \multicolumn{1}{l|}{\textbf{Dust}} & 36.89 & 22.37 & 8.65 & 16.14 & 21.01 & 0 & 0 & \multicolumn{1}{c|}{14.01} & 57.64 & - & 77.3 & 78.8 & 71.25 & - & - & 71.25 \\
\multicolumn{1}{l|}{\multirow{4}{*}{\textbf{\begin{tabular}[c]{@{}l@{}}FasterRCNN \\ MobileNet   \\ Large\\ \protect\cite{ren2015faster,howard2017mobilenets}\end{tabular}}}} & \multicolumn{1}{l|}{\textbf{Rain}} & 63.1 & 50.16 & 12.41 & 46.25 & 42.98 & 0 & - & \multicolumn{1}{c|}{34.38} & 51.34 & 50.4 & 73.85 & 91.43 & 66.76 & 0 & - & 53.41 \\
\multicolumn{1}{l|}{} & \multicolumn{1}{l|}{\textbf{Snow}} & 58.96 & 39.59 & 7.6 & 40.27 & 36.60 & 0 & 0 & \multicolumn{1}{c|}{20.92} & 52.03 & 43.02 & 30 & 66.32 & 47.84 & 0 & - & 38.27 \\
\multicolumn{1}{l|}{} & \multicolumn{1}{l|}{\textbf{Fog}} & 63.41 & 20.83 & 29.2 & 34.06 & 36.87 & 0 & 0 & \multicolumn{1}{c|}{24.58} & 55.84 & \textbf{100} & 81.42 & 75.03 & 78.07 & \textbf{100} & 0 & 68.72 \\
\multicolumn{1}{l|}{} & \multicolumn{1}{l|}{\textbf{Dust}} & 58.28 & 33.15 & 14.63 & 20.16 & 31.55 & 0 & 0 & \multicolumn{1}{c|}{21.04} & \textbf{81.64} & - & 79.08 & 86.73 & 82.48 & \textbf{-} & - & 82.49 \\
\multicolumn{1}{l|}{\multirow{4}{*}{\textbf{\begin{tabular}[c]{@{}l@{}}FasterRCNN \\ ResNet 50\\ \protect\cite{ren2015faster}\end{tabular}}}} & \multicolumn{1}{l|}{\textbf{Rain}} & 72 & \textbf{71.46} & 23.1 & \textbf{50.01} & \textbf{54.14} & 0 & - & \multicolumn{1}{c|}{\textbf{43.31}} & 53.11 & 57.55 & 73.74 & \textbf{92} & 69.1 & 0 & - & 55.28 \\
\multicolumn{1}{l|}{} & \multicolumn{1}{l|}{\textbf{Snow}} & 66.13 & 58.97 & 27.61 & 40.38 & 48.27 & 0 & 0 & \multicolumn{1}{c|}{27.58} & 53.17 & 38.74 & 40.74 & 67.53 & 50.05 & 0 & - & 40.03 \\
\multicolumn{1}{l|}{} & \multicolumn{1}{l|}{\textbf{Fog}} & \textbf{72.53} & 45.11 & \textbf{33.08} & 31.8 & 45.63 & 0 & 0 & \multicolumn{1}{c|}{30.42} & 59.07 & \textbf{100} & \textbf{82.77} & 76.01 & 79.46 & \textbf{100} & 0 & 69.64 \\
\multicolumn{1}{l|}{} & \multicolumn{1}{l|}{\textbf{Dust}} & 66.68 & 51.99 & 30.28 & 26.24 & 43.79 & \textbf{0.7} & 0 & \multicolumn{1}{c|}{29.31} & 81.35 & - & 80.37 & 87.16 & \textbf{82.96} & - & - & \textbf{82.96} \\ \hline
\end{tabular}}

\caption{\textbf{Object Detection broken down by weather: } 
We present results for DAWN-All and WEDGE across four weather conditions: rain (storm), snow (storm), fog, dust (haze, mist, sand). Note that some objects occur rarely in certain weather conditions (e.g., people bike less during storms), resulting in performance estimates that may not be as reliable. Additionally, the number of samples across different weather conditions is unbalanced which causes certain weather conditions to impact overall detection scores differently. Looking at average metrics across object categories, good-weather-trained models generalize better to snow and fog compared to rain and dust. Fine-tuning trucks on WEDGE consistently improves real-world performance across most weather conditions.}

\label{tab4}
\end{center}

\end{table*}

\subsection{Classification Benchmark}

As visible in Table \ref{tab2}, the MobileNet \protect\cite{howard2017mobilenets} Classifier achieves top performance on the WEDGE Dataset with 53.87\% test accuracy which is over 8-fold improvement on random classification that hits 6.25\% accuracy.

\subsection{Object Detection Benchmark}

The main task of this study is examining WEDGE's usefulness in robust object detection across multi-weather adversarial environments. We focus our results on the real-world DAWN dataset\protect\cite{kenk2020dawn}. Previous work uses different protocols for evaluation on DAWN; \protect\cite{ms} evaluates on DAWN WD set (Fake droplets on fake wet generated conditions) and reports the overall AP averaged over classes, \protect\cite{saikia2021improving} evaluates on corrupted testsets and reports average AP across corruptions, \protect\cite{marathe2022rain,marathe2022restorex} evaluates on 1000 random images, while\protect\cite{a3} evaluates on 500 random images of DAWN and reports AP and mAP. Previous work \protect\cite{rs} evaluates on a 3:1 train-test holdout split of DAWN (trained on adverse weather data) and reports overall Vehicle AP, including comparisons to previous models \cite {cai2018cascade,hu2018sinet,hassaballah2020local,chabot2017deep,xiang2017subcategory,lin2017focal,redmon2018yolov3,li2019scale, he2017mask} in classic supervised learning settings.  
 DAWN has a (proposed) 90-10 train-test split, but since our models are not trained on DAWN, we present results for both DAWN-All (Table \ref{tab3}) and DAWN-Test (Table \ref{tab6}).

{\bf Off-the-shelf (OTS).} First, we find that simply evaluating state-of-the-art (SOTA) {\em off-the-shelf} (OTS) object detectors (trained on {\em good} weather data) already outperforms all published results. Because prior work often evaluated with different protocols that complicate comparisons, we first establish a standard DAWN benchmark, obtaining dramatically better performance on 4 common categories (T-4) with 17.07 T-4 AP increase (on DAWN-All) and 22.97 AP increase (DAWN-Test), compared to previous state-of-the-art ensemble methods\protect\cite{a3} (AP@50 reported in some works is included in brackets: Improvement over this value is 12.96 AP on DAWN-All and 18.86 AP on DAWN-Test). This establishes our pre-trained detectors as strong baselines for this task. It is important to re-emphasize that many previous works (Table \ref{tab3}) evaluate on different DAWN test sets, making strict comparisons difficult. To enable consistent future evaluations, we will publish our DAWN split (inclusive of all 4 weather conditions). 

{\bf Fine-tuning.} Second, fine-tuning such models (specifically, ResNet50) on WEDGE further improves truck AP by 4.48 on DAWN-All (4.44 AP on DAWN-Test). The fine-tuned MobileNet-Large is able to detect both cars and trucks better with  1.96 AP and 9.17 AP on DAWN-All and (2.61 AP and 5.17 AP on DAWN-Test) respectively.  
In general, we find that fine-tuning hurts performance for other categories, suggesting that it may be useful to explore additional fine-tuning data or architectures that learn without forgetting pre-trained knowledge. 

{\bf WEDGE Benchmark.} Third, we also evaluate on synthetic WEDGE data in Table \ref{tab3}. The best object detection models in classical supervised settings (fine-tuned and tested on WEDGE) attaining 45.41 mAP on WEDGE, with highest AP 57.48  for {\tt car} obtained by Faster-RCNN (ResNet50). 

{\bf Weather Analysis.} Fourth, we analyze object detection performance under each weather category provided in DAWN, as presented in Table \ref{tab4}. The overall trends mirror the observations from Table \ref{tab3}, in the sense that OTS detectors work well across all weather conditions and fine-tuning on WEDGE improves truck AP by good margins with an averaged 9 AP improvement. {\tt Fog} results in higher detection scores with the OTS detectors reaching 62.7 mAP, while {\tt dust} is most challenging with the lowest mAP score of 23.21. When evaluated on synthetic data (WEDGE), {\tt dust} produces higher detection scores with the best OTS detectors reaching 41.45 mAP and best fine-tuned models reaching 82.96 mAP. This behaviour may be indicative of a Sim2Real gap in weather simulation. 
In synthetic data settings (WEDGE), {\tt snow} is most challenging, yielding the lowest mAP of 15.06.
Overall, it is important to acknowledge that fine-tuning on WEDGE seems most effective for object classes that are well-generated with a low Sim2Real gap (trucks), but this does not hold consistently for other object categories. In the next section, we manually examine how the synthetic objects in these class are significantly worse than real images which cause the detector to fine-tune on incorrect representations, thus hampering performance.

\section{Discussion}

{\bf Qualitative analysis of WEDGE.} As shown in Fig. \ref{fig10}, we conduct qualitative analysis on generated samples and summarize our qualitative observations. {\tt Snow} class examples closely resemble winter scenes which contain noisy elements like snowfall and thus poor visibility conditions follow.  {\tt Rain} examples resemble the view of a rainy traffic-filled road from the perspective of a sensor placed behind windshield. {\tt Dust} contains occluded objects which are annotated for robust vision in adversity. {\tt Fog} examples showcase dense foggy conditions which impair visibility of pedestrians and objects. {\tt Sun} imagery have well-illuminated objects in variety of backgrounds. {\tt Lightning} images look realistic but typically contain a higher proportion of sky pixels. {\tt Cloudy} examples resemble true cloudy scenarios with reduced illumination and gray overcasts.  {\tt Hurricane} consists of images that appear un-realistic, likely to due to the fact that this extreme weather condition is relatively rare. {\tt Night} images have poor illumination and make detection difficult as expected. Often distant vehicles are just shown by blurred lights which we have included in annotations to ensure that vehicles (detectors) can even detect distant mobile objects under low illumination. {\tt Summer} are generally well-lit images. {\tt Spring} images appear difficult to differenciate from day and sun, which is favorable as spring is a transitional season. {\tt Winter} contains elements like snow, blizzards, hail which heavily obstruct vision and provide good adversaries to the detection task. Backgrounds are mostly white and snow–covered which makes the detection task simpler. This does not represent winter in warmer countries, which must be treated by mixing classes. {\tt Fall} images are skewed to geographic regions that are usually associated with the aesthetic fall backgrounds including bright trees, fallen leaves which are mostly present in the northern regions of countries. {\tt Tornado} contains a good number of unrealistic images as well, but manages to capture the essence of this natural disaster through poor-illumination, windy conditions and distant tornado funnels. In the unrealistic cases, tornadoes appear in extremely unlikely scenarios like exactly on top of the car, as visualized in cartoons and games. {\tt Day} images are well-lit and show sunny scenarios, also including some overcast skies. {\tt Windy} images are either realistic or extremely skewed towards disaster-like scenarios including uprooting winds, destroyed vehicles and fading objects.

\begin{figure}[h!]
    \centering    \includegraphics[width=0.49\textwidth,height=9cm]{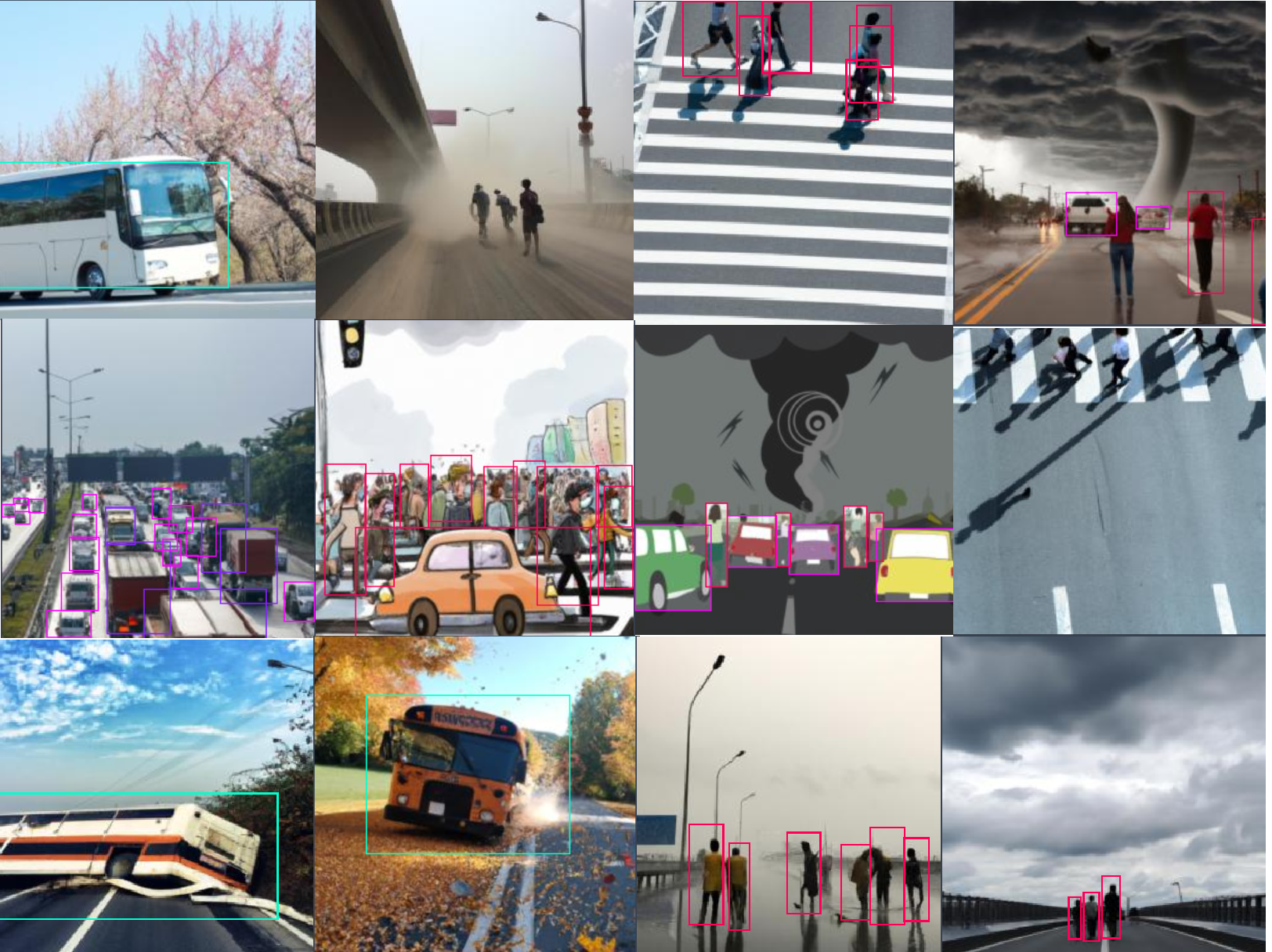}
    \caption{\textbf{Qualitative Analysis: }Limitations of WEDGE appear in the form of  region-centric spurious correlations, generative anomalies,missing (incomplete generation) features, domain and style transfer, distance (proximity to viewing angle) bias, multi-time relevance, class (weather) bias, spatial and placement anomalies,human generation inconsistencies from top-left to bottom-right.}
    \label{fig9}
\end{figure}

{\bf Anomalies observed in WEDGE.} As visible in Fig. \ref{fig9} we highlight some possible causes of poor performance. Region-centric correlations (eg. cherry blossoms associated with spring), are a recurring theme in the generated images inspite of providing generic prompts. Generative anomalies like extra terrestrial creatures crossing the road occur when the terrain described in prompt (dust) matches similar out-of-distribution examples (Martian imagery). Training objects sometimes combine to form interesting but unrealistic characters in this synthetic in spite of given realistic prompts. We also identify entities with incomplete generations or missing parts. While this feature can help improve robustness to occlusion, it is still a limitation of generated images. Often typical scenes which correspond to the prompt are generated in sketch, animated or miscellaneous styles. Objects which are closer to the viewer (camera)’s supposed location are more accurately generated. As seen in the figure and other examples, the distant objects are often lacking quality and fundamental differentiating characteristics which are necessary for detectors. Although we cannot accurately pinpoint the time frame of generated images, we observe special cases of people wearing masks in locations (predicted locations) where masks were not worn prior to the pandemic. While this may be attributed to different reasons, we can consider this feature as an important part of robustness in post-pandemic systems.  As the prompts shift to more out-of-distribution settings, like tornadoes, we observe a dramatic shift in favor of unrealistic images. This may be due to the inability to find hyper-realistic training images captured in these adverse conditions, but are a potential limitation. Spatial anomalies  frequently distort the placement, positioning, orientation and interaction between generated objects. In this case, we observe shadows are generated inconsistently. As generative models move closer towards real-world simulation, focus on modeling relationships between entities on the basis of physical, scientific and behavioral properties can be explored. Beneficial anomalies like scenes generated around accidents, mishaps like tire punctures, car crashes and weather-related disasters like tornadoes uprooting the roofs of buses appear often in the data. These accidents are very realistic and not often captured by common autonomous vehicle datasets. These scene-specific datasets can be generated for detecting emergencies in  surveillance systems. Human generation ultimately presents the greatest challenge in dataset utilization. The images of humans in the dataset have second-largest frequency but are often unrealistic (either due to the out-of-distribution prompts or intentional obscuring done for privacy concerns) which can potentially affect fine-tuning as seen in the previous section.

{\bf Benefits of generated datasets (WEDGE).} The feature and capability of embedding variability from vast text corpora into image sets using prompting of generative models provides support for building robust models. Due to the high variability in geography, population, seasons, weather, illumination, perspective and backgrounds, models are able to generalize to detect trucks on real roads. We can simulate specific out-of-distribution scenarios like road accidents to monitor safety through anomaly detection or other tasks.

{\bf When does WEDGE work?} The image generation procedure and results of this study speak in favor of its importance to the autonomous driving perception tasks. Prompting was focused on generating the most relevant autonomous vehicle-related images for 16 weather-classes and manually verified. Image screening and curation was performed to ensure inter-class-prompt consistency. The provided annotations and extensive bounding boxes (16513) for all classes have been generated with human-in-the-loop.  16 unique weather-seasonal variations captured for autonomous vehicles which is unique to this dataset and essential for multi-weather robustness. Annotations for heavily occluded and obscure objects (headlights in fog) have been labelled to assist models in learning representations from occluded objects. Inspite of having out-of-distribution scenarios, the image similarity thresholds are still within reasonable range from the sample distribution shifts which speaks in favor of data adoption in similar tasks needing sensor-based data. Models trained on WEDGE for domain adaptive detection were able to cross the benchmark on the DAWN dataset in under-represented target classes like trucks. The difference between generated people and trucks and their similarities to real-world objects which differ dramatically between real and synthetic data, offer a plausible explanation for the performance difference. 

\begin{figure}[h]

        \centering
       \includegraphics[width=\columnwidth,height=5cm]{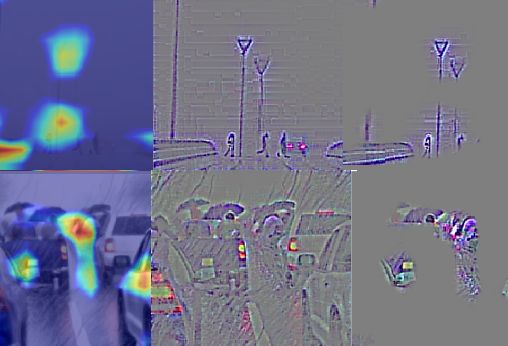}
    \caption{\textbf{WEDGE as an adversarial example:} We observe significant shifts in attention maps\protect\cite{selvaraju2017grad} when data contains poor-weather conditions. The object of interest was the vehicle in the images which the attention maps are not following due to the weather-based corruptions of fog and dust. This provides support to why good-weather data are often insufficient while building robust perception models.}
    \label{fig7}

\end{figure}

\begin{figure}[h]

        \centering
       \includegraphics[width=\columnwidth,height=9cm]{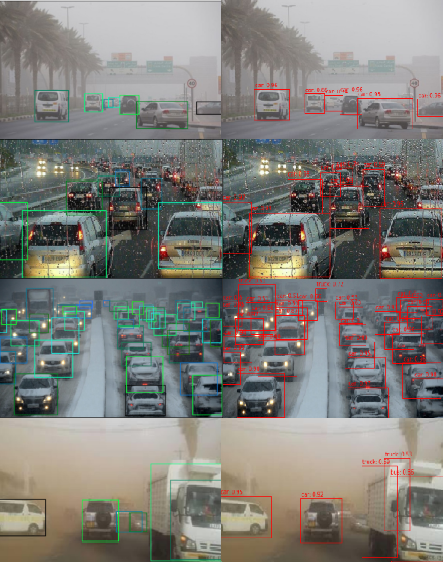}
    \caption{\textbf{Sim2Real Inference: }Comparison of (COCO) pre-trained Resnet 50 Faster RCNN ({\bf left}) with a variant fine-tuned on WEDGE ({\bf right}) on a test image from DAWN. We see that the fine-tuned models tend to predict trucks better but suffer from false positives, resulting in lower car APs.}
    \label{fig8}

\end{figure}

\section{Conclusion}

In this work, we explore AI-generated datasets\footnote{All references to "generated" in this text imply AI generated datasets only. The authors generated this dataset in part with DALLE-2, OpenAI’s large-scale image-generation model. Upon generating the dataset, the authors reviewed the images and take responsibility for their content in accordance with the terms laid out by OpenAI. The authors have created "Input" prompts on their own and obtained data "Output" images \textbf{only} using the official OpenAI API through a paid subscription service.} for robust multi-weather perception. We perform a small-scale analysis of its task-specific properties in the context of autonomous vision and demonstrate the selective effectiveness of such generation. Under the constraints of selected data, we assess the usefulness of these datasets from the perspective of autonomous perception. We acknowledge that all findings are constrained to this case study between the selected domain and target data only, and do not present findings for autonomous perception or synthetic data in general. 

In this work, we additionally present a state-of-the-art benchmark for DAWN dataset using standard evaluation metrics and OTS detectors (without any access to target or adverse-weather training data). We hope to aid in the effort towards meeting the need for autonomous vision datasets by this demonstration. 

In the development of safe autonomous systems and robust perception models, all-weather vision should be an important consideration. The corruptions introduced by adverse weather may differ in real and synthetic datasets, depending on the nature of the selected weather and the method of image generation. Bridging the Sim2Real gap in weather simulation, particularly for out-of-distribution weather scenarios like tornadoes was highlighted in WEDGE. Once bridged, models that perform robustly under adverse weather corruptions can be tested with realistic prompt-driven synthetic adversarial examples. In future works, this data generation procedure paired with creative prompt engineering can work towards delivering superior performance in multi-weather domains.

{\small
\bibliographystyle{ieee_fullname}
\bibliography{egbib}
}
\clearpage

\appendix
\section{Appendix}

\subsection{Extended Results}

We present comparisons across the proposed DAWN-Test and DAWN-All in Table \ref{tab6}.

\begin{table*}[htbp!]
\begin{center}

\resizebox{1.9\columnwidth}{!}{\begin{tabular}{lllllllllllllll}
\hline
\multicolumn{1}{l|}{} & \multicolumn{8}{l|}{\textbf{Real Data (DAWN Dataset)}} & \multicolumn{6}{l}{\textbf{Synthetic Data (WEDGE Dataset)}} \\ \cline{2-15} 
\multicolumn{1}{l|}{\multirow{-2}{*}{\textbf{Model}}} & \textbf{car} & \textbf{person} & \textbf{bus} & \textbf{truck} & \textbf{T-4 AP} & \textbf{mc} & \textbf{bicycle} & \multicolumn{1}{l|}{\textbf{mAP}} & \textbf{car} & \textbf{person} & \textbf{bus} & \textbf{truck} & \textbf{van} & \textbf{mAP} \\ \hline
\multicolumn{15}{l}{\textbf{Prior Art}} \\ \hline
\multicolumn{1}{l|}{Multi-weather city \protect\cite{ms}} & - & - & - & - & 21.20 (39.19) & - & - & \multicolumn{1}{l|}{(39.19)} & - & - & - & - & - & - \\
\multicolumn{1}{l|}{RoHL \protect\cite{saikia2021improving}} & - & - & - & - & - & - & - & \multicolumn{1}{l|}{28.80} & - & - & - & - & - & - \\
\multicolumn{1}{l|}{Transfer Learning \protect\cite{marathe2022rain}} & 7.00 & 8.00 & 7.00 & - & 5.50 & - & 0.00 &  \multicolumn{1}{l|}{-} & - & - & - & - & - & - \\
\multicolumn{1}{l|}{Data Augmentation \protect\cite{marathe2022rain}} & 6.00 & 4.00 & 3.00 & 0.00 & 26.25 & - & \textbf{92.00} & \multicolumn{1}{l|}{-} & - & - & - & - & - & - \\
\multicolumn{1}{l|}{\begin{tabular}[c]{@{}l@{}}Weather-\\ Night GAN \protect\cite{marathe2022restorex}\end{tabular}} & 48.00 & 0.00 & 0.00 & 0.00 & 12.00 & - & - & \multicolumn{1}{l|}{-} & - & - & - & - & - & - \\
\multicolumn{1}{l|}{Ensemble Detectors \protect\cite{a3}} & 52.56 & 52.34 & 21.73 & 13.71 & 35.08 & 35.51 & 23.29 & \multicolumn{1}{l|}{32.75} & - & - & - & - & - & - \\ \hline \hline
\multicolumn{15}{l}{\textbf{Evaluation on DAWN-All}} \\ \hline
\multicolumn{15}{l}{\textbf{Trained on Good Weather Data (COCO \protect\cite{lin2014microsoft})}} \\ \hline

\multicolumn{1}{l|}{\begin{tabular}[c]{@{}l@{}}FasterRCNN   \\ MobileNet \\ Large 320 \protect\cite{ren2015faster,howard2017mobilenets}\end{tabular}} &37.56	&	34.93	&	20.90	&	12.91	&	26.57	&	23.15	&	18.95	&	\multicolumn{1}{l|}{24.73} & 34.10	&	36.26	&	39.35	&	16.05	&	0.00	&	25.15\\
\multicolumn{1}{l|}{\begin{tabular}[c]{@{}l@{}}FasterRCNN   \\ MobileNet\\ Large \protect\cite{ren2015faster,howard2017mobilenets}\end{tabular}} & 60.64	&	55.96	&	32.78	&	23.66	&	43.26	&	38.55	&	28.75	&	\multicolumn{1}{l|}{40.05} & 35.34	&	39.52	&	35.83	&	25.43	&	0.00	&	27.22\\
\multicolumn{1}{l|}{FasterRCNN ResNet 50\protect\cite{ren2015faster}} &  \textbf{69.13}	&	\textbf{70.31}	&	\textbf{38.64}	&	30.54	&	\textbf{52.15}	&	\textbf{52.17}	&	\textbf{30.56}	&	\multicolumn{1}{l|}{\textbf{48.55}}  & 31.41	&	33.54	&	30.19	&	18.75	&	0.00	&	22.78 \\ \hline

\multicolumn{15}{l}{\textbf{Fine-Tuning on WEDGE}} \\ \hline
\multicolumn{1}{l|}{\begin{tabular}[c]{@{}l@{}}FasterRCNN   \\ MobileNet \\ Large 320 \protect\cite{ren2015faster,howard2017mobilenets}\end{tabular}} & \textbf{39.52}	&	23.97	&	7.81	&	\textbf{22.08}	&	23.34	&	0.00	&	0.00	&	\multicolumn{1}{l|}{15.56} & 40.40	&	43.01	&	49.88	&	31.41	&	10.19	&	34.98 \\
\multicolumn{1}{l|}{\begin{tabular}[c]{@{}l@{}}FasterRCNN   \\ MobileNet \\ Large \protect\cite{ren2015faster,howard2017mobilenets}\end{tabular}} & 59.81	&	34.61	&	14.06	&	\textbf{30.67}	&	34.78	&	0.00	&	0.00	&	\multicolumn{1}{l|}{23.19}  & 52.52	&	\textbf{54.79}	&	\textbf{51.23}	&	50.01	&	7.95	&	43.30\\ 
\multicolumn{1}{l|}{FasterRCNN ResNet 50\protect\cite{ren2015faster}} &  68.09	&	54.29	&	27.48	&	\textbf{35.02}	&	46.22	&	0.00	&	0.00	&	\multicolumn{1}{l|}{30.81} & \textbf{57.48}	&	54.71	&	46.92	&	\textbf{57.43}	&	\textbf{10.49}	&	\textbf{45.41}\\ \hline
\multicolumn{15}{l}{\textbf{Evaluation on DAWN-Test}} \\ \hline
\multicolumn{15}{l}{\textbf{Trained on Good Weather Data (COCO \protect\cite{lin2014microsoft})}} \\ \hline
\multicolumn{1}{l|}{\begin{tabular}[c]{@{}l@{}}FasterRCNN   \\ MobileNet \\ Large 320 \protect\cite{ren2015faster,howard2017mobilenets}\end{tabular}} & 39.08 & 22.71 & 37.13 & 10.78 & 27.42 & 8.33 & 0.00 & \multicolumn{1}{l|}{19.70} & 34.10	&	36.26	&	39.35	&	16.05	&	0.00	&	25.15
 \\
\multicolumn{1}{l|}{\begin{tabular}[c]{@{}l@{}}FasterRCNN   \\ MobileNet \\ Large \protect\cite{ren2015faster,howard2017mobilenets}\end{tabular}} & 60.26 & 36.74 & 49.30 & 17.94 & 41.06 & 23.33 & 0.00 & \multicolumn{1}{l|}{31.26} & 35.34	&	39.52	&	35.83	&	25.43	&	0.00	&	27.22 \\
\multicolumn{1}{l|}{FasterRCNN ResNet 50\protect\cite{ren2015faster}} & \textbf{71.19} & \textbf{69.51} & \textbf{69.88} & 21.62 & \textbf{58.05} & 25.00 & 20.00 & \multicolumn{1}{l|}{\textbf{46.20}} & 31.41	&	33.54	&	30.19	&	18.75	&	0.00	&	22.78 \\ \hline
\multicolumn{15}{l}{\textbf{Fine-Tuning on WEDGE}} \\ \hline
\multicolumn{1}{l|}{\begin{tabular}[c]{@{}l@{}}FasterRCNN   \\ MobileNet \\ Large 320 \protect\cite{ren2015faster,howard2017mobilenets}\end{tabular}} & \textbf{41.69} & 19.02 & 16.79 & \textbf{15.95} & 23.36 & 0.00 & 0.00 & \multicolumn{1}{l|}{15.57} & 40.40	&	43.01	&	49.88	&	31.41	&	10.19	&	34.98\\
\multicolumn{1}{l|}{\begin{tabular}[c]{@{}l@{}}FasterRCNN   \\ MobileNet\\ Large \protect\cite{ren2015faster,howard2017mobilenets}\end{tabular}} & 58.54 & 28.39 & 29.14 & \textbf{21.68} & 34.43 & 0.00 & 0.00 & \multicolumn{1}{l|}{22.96} & 52.52	&	\textbf{54.79}	&	\textbf{51.23}	&	50.01	&	7.95	&	43.30
 \\
\multicolumn{1}{l|}{FasterRCNN ResNet 50\protect\cite{ren2015faster}} & 65.47 & 39.70 & 54.19 & \textbf{26.06} & 46.35 & 0.00 & 0.00 & \multicolumn{1}{l|}{30.9} & \textbf{57.48}	&	54.71	&	46.92	&	\textbf{57.43}	&	\textbf{10.49}	&	\textbf{45.41}  \\ \hline \hline
\end{tabular}}
\caption{\textbf{Object Detection: } Performance for {\tt Car, Person, Bus, Truck, Van, Motorcycle (mc), Bicycle} using the PASCAL VOC mAP metric on real (DAWN) and our synthetic (WEDGE) data. Previous work uses different protocols for evaluation on DAWN; \protect\cite{ms} evaluates on DAWN WD set (Fake droplets on fake wet generated conditions) and reports the overall AP averaged over classes (AP @50 is included in brackets: Improvement over this value is 12.96 AP on DAWN-All and 18.86 AP on DAWN-Test), \protect\cite{saikia2021improving} evaluates on corrupted testsets and reports average AP across corruptions, \protect\cite{marathe2022rain,marathe2022restorex} evaluates on 1000 random images, while\protect\cite{a3} evaluates on 500 random images of DAWN and reports AP and mAP. DAWN has a proposed 90-10 train-test split, but since our models are not trained on DAWN, we present results for both DAWN-Test and DAWN-All. First, we find that simply evaluating state-of-the-art (SOTA) {\em off-the-shelf} (OTS) object detectors (trained on {\em good} weather data) already outperforms all published results. This establishes our pre-trained detectors as strong baselines for this task. Fine-tuning such models (specifically, ResNet50) on WEDGE further improves truck AP by 4.44 AP on DAWN-Test (4.48 on DAWN-All). The fine-tuned MobileNet-Large is able to detect both cars and trucks better with 2.61 AP and 5.17 AP on DAWN-Test and (1.96 AP and 9.17 AP on DAWN-All) respectively. T-4 AP is the averaged AP over 4 key object classes {\tt Car, Person, Bus, Truck}.}
 \label{tab6}
     
\end{center}
\end{table*}

\end{document}